# Advancing Italian Biomedical Information Extraction with Transformers-based Models: Methodological Insights and Multicenter Practical Application

Claudio Crema[a, 1] [0000-0003-2537-9742], Tommaso Mario Buonocore[b] [0000-0002-2887-088X], Silvia Fostinelli[c] [0000-0003-0120-880X], Enea Parimbelli[b] [0000-0003-0679-828X], Federico Verde[d, e] [0000-0002-3977-6995], Cira Fundarò[f] [0000-0002-7810-8885], Marina Manera[g] [0000-0003-4235-9612], Matteo Cotta Ramusino[h] [0000-0003-3090-9648], Marco Capelli[h] [0000-0002-4454-366X], Alfredo Costa[h] [0000-0001-7312-8011], Giuliano Binetti[c] [0000-0003-2759-5844], Riccardo Bellazzi[b] [0000-0002-6974-9808], Alberto Redolfi[a] [0000-0002-4145-9059]

a: Laboratory of Neuroinformatics, c: Molecular Markers Laboratory, IRCCS Istituto Centro San Giovanni di Dio Fatebenefratelli, Brescia, Italy
{ccrema, sfostinelli, gbinetti, aredolfi}@fatebenefratelli.eu

b: Dept. of Electrical, Computer and Biomedical Engineering, University of Pavia, Pavia, Italy
buonocore.tms@gmail.com
{riccardo.bellazzi, enea.parimbelli}@unipv.it

d: Department of Neurology and Laboratory of Neuroscience, IRCCS Istituto Auxologico Italiano, Milan, Italy
e: Department of Pathophysiology and Transplantation, Dino Ferrari Center, Università degli Studi di Milano, Milan, Italy
f.verde@auxologico.it

f: Neurophysiopatology Unit, g: Psychology Unit, IRCCS Istituti Clinici Scientifici Maugeri, Pavia Italy
{cira.fundaro, marina.manera}@icsmaugeri.it

h: Unit of Behavioral Neurology, IRCCS Mondino Foundation Pavia, and Dept. of Brain and Behavioral Sciences, University of Pavia, Pavia, Italy
{alfredo.costa, marco.capelli, matteo.cottaramusino}@mondino.it

**Abstract.** The introduction of computerized medical records in hospitals has reduced burdensome activities like manual writing and information fetching. However, the data contained in medical records are still far underutilized, primarily because extracting data from unstructured textual medical records takes time and effort. Information Extraction, a subfield of Natural Language Processing, can help clinical practitioners overcome this limitation by using automated text-mining pipelines. In this work, we created the first Italian neuropsychiatric Named Entity Recognition dataset, PsyNIT, and used it to develop a Transformers-based model. Moreover, we collected and leveraged three external independent datasets to implement an effective multicenter model, with overall F1-score 84.77%, Precision 83.16%, Recall 86.44%. The lessons learned are: (i) the crucial role of a consistent annotation process and (ii) a fine-tuning strategy that combines classical methods with a "low-resource" approach. This allowed us to establish methodological guidelines that pave the way for Natural Language Processing studies in less-resourced languages.

**Correspondence:** ccrema@fatebenefratelli.eu

# Introduction

The ubiquity of digital technologies is increasingly encompassing every aspect of our lives, and healthcare is no exception. In the last years there has been a rapid adoption of digital health tools [1]. This new technological paradigm has led to a dramatic increase in digitized medical text data in the everyday medical routine of healthcare institutions (e.g., discharge letters, examination results, medical notes) [2]. These documents, while very informative, are unstructured and not harmonized, creating a barrier that leads to insufficient use and under-exploitation. This lowers the efficiency of the clinical and research environments, since the extraction of such information into structured databases is time-consuming: physicians spend about 35% of their time documenting patient data [3].

Artificial Intelligence (AI), and in particular Natural Language Processing (NLP), could provide useful tools to overcome these limitations. NLP is a collection of techniques and tools for processing human language written texts. Some examples of NLP tasks are: Named Entity Recognition (NER), which assigns words to predefined categories (e.g., person, location); Relation Extraction (RE), which connects named entities in a text through semantic relations; and Question Answering (QA), whose goal is to find answers to questions written by humans. In the last decade, NLP has shifted to Deep Learning (DL) approaches, and a large number of models have been implemented. The advent of the Transformer architecture [4] unlocked the creation of highly performing models, and the famous Bidirectional Encoder Representations from Transformers (BERT) [5], developed by Google in 2019, established itself as the de-facto state-of-the-art. Several BERT-based models followed shortly after. These models are usually created in a two-step process:

- The first step is *pre-training*, an unsupervised procedure in which the model is fed with a huge amount of unlabeled text (e.g., the BERT corpus is composed of 3.3 billion words). The pre-training is based on the mechanism of Masked Language Modeling (MLM): a random portion of the words in a sentence is masked, and the model tries to predict them based on the surrounding context. At the end of the pre-training, the model has a general knowledge of the language. Texts used for pre-training are usually referred to as corpora (i.e., a large collections of written texts).

- The second step is *fine-tuning*, a supervised training in which the model is fed with a relatively small set of labeled training examples (e.g., a famous QA dataset, the Stanford Question Answering Dataset (SQuAD) [6], is composed of 100 thousand examples), and learns to perform a specific task. When speaking about fine-tuning, we usually refer to datasets, which are a structured collection of data used for a specific purpose.

One of the main limitations of this process is that it requires a considerable amount of text in the pre-training phase to achieve good results. For this reason, the models available in literature are often trained on generic corpora (e.g., BERT main corpus is the English Wikipedia), and they have difficulties when it comes to specific topics.

However, efforts have been made to overcome this limitation. Biomedical BERT (BioBERT [7]), is one of the best known and most successful models. This model was developed using the same approach as the original BERT, with the key difference that the pre-training corpus consists of PubMed abstracts and full-text articles, totaling 18 billion words. BioBERT performs better than the original BERT when applied to various NLP tasks involving biomedical documents. This result proves that the use of a topic-specific pre-training corpus is a crucial factor for high performance in a specific domain such as biomedicine. Another example is SciBERT [18], which exploits the original BERT architecture but trains the model from scratch on a different dataset (1.14 million scientific papers from Semantic Scholar). Thanks to this, SciBERT is able to incorporate a custom dictionary that reflects the in-domain word distribution more accurately. Along this path, several biomedical NLP models have recently been proposed to address the aforementioned NLP tasks:

- BioNER [8, 9], used to identify specific medical entities in a text (e.g., drugs, medical tests, dosages, scores). Dihn et al. [19] developed a tool able to identify antibody and antigen entities and achieved a F1-score of 81.44%. Li et al. [21] compared four biomedical BERT models (BioBERT, SciBERT, BlueBERT [22], PubMedBERT [23]) and two open-domain models (BERT and SpanBERT [24]) by fine-tuning them on three clinical datasets, showing that the domain-specific models outperformed the open-domain ones with the best model achieving an F1-score of 83.6%. Yeung et al [47] developed a BioBERT-based tool to identify metabolites in cancer-related metabolomics articles with an F1-score of 90.9%. Dang et al. [48] created a Long Short-Term Memory (LSTM [49]) network and tested it on the NCBI disease dataset [40] with an F1-score of 84.41%, while Cho et al. [50] performed the same test with their bi-direction LSTM-based tool and achieved an F1-score of 85.68%. Finally, it is worth noting that although the vast majority of these tools were developed using English corpora, some work has been done for other languages as well. Chen et al. [51] developed a BERT-based hybrid network, and Li et al [52] developed a DL model incorporating dictionary features. These systems were tested on the China Conference on Knowledge Graph and Semantic Computing dataset 2017 version, and achieved F1-scores of 94.22% and 91.60%, respectively.

- BioRE [10, 11], used after NER, in order to connect medical entities (e.g., drugs and their dosages).

- BioQA [12, 13], aimed at finding answers to specific questions in a medical text.

BioNER, BioRE, and BioQA are tools used in Information Extraction (IE), one of the NLP main subtasks. IE has the goal of making the semantic structure of a text explicit, so that we can make use of it [14]; an example is available in Supplementary notes, in the section "Information Extraction example".

Some missing points need to be highlighted. The so-called "less-resourced languages", e.g., Italian, are underrepresented in this scenario. Indeed, models for very specific medical topics in these languages are lacking, although some examples can be found in literature ([15, 16]). This is due to the fact that, also for the biomedical topic, the vast majority of models are trained on English corpora, mainly because it is difficult to find a sufficiently large medical corpus in these languages [17].

In this paper we try to overcome these limitations by using Italian biomedical BERT models and fine-tuning them for the NER task on a specific medical topic, namely neuropsychiatry. The models we have created could be used to implement IE tools, avoiding lengthy and repetitive procedures by highly specialized clinical staff.*Statement of significance*

| | |
|---|---|
| **Problem** | Clinical reports are highly informative, but heavily underused because data within them is burdensome to extract. |
| **What is already known** | NER tools could help to automatize the extraction process. However, there is a lack of tools for very specific topics (e.g., neuropsychiatry) in less-resourced languages (e.g., Italian). |
| **What this paper adds** | This study creates an effective Italian neuropsychiatry multicenter NER model and share it with the community, along with one of the NER datasets used. We conducted several experiments and analized the multicenter fine-tuning process, proposing methodological guidelines that we believe can be applied to other topics for less resourced languages as well. |

# Objectives

This work makes the following contributions:

A. A native Italian neuropsychiatric NER dataset, called PsyNIT (Psychiatric Ner for ITalian), with about six thousand entities, used to fine-tune an Italian biomedical NER model. To our best knowledge, these are the first less-resourced dataset[1] and model[2] publicly available for neuropsychiatry.

B. It further validates the checkpoints developed in [25] on a specific NER downstream task that is both on natively Italian text and clinically relevant in the neuropsychiatry domain.

---

[1] https://huggingface.co/datasets/Neuroinformatica/PsyNIT
[2] https://huggingface.co/IVN-RIN/MedPsyNIT

C. It examines in detail the process of implementing a biomedical multicenter NER model and provides methodological guidance for the model development. We believe that these guidelines are generalizable to other domains as well.

# Methods

## Pre-training

The pre-training of the starting Italian biomedical checkpoints followed the procedure described in [25]. The original model was a general-purpose Italian BERT checkpoint[3] (called BaseBIT). The pre-training corpus is a composition of an Italian Wikipedia dump, OPUS, and OSCAR corpora, for a total size of 81 GB and 13 billion tokens. Following the original BioBERT approach, this checkpoint was further pre-trained with a biomedical corpus, obtained by the automatic translation of the original BioBERT corpus (over 2 million PubMed abstracts), for a total size of 28 GB. The translation was carried out by leveraging Google neural machine translation, an RNN-based framework. This checkpoint was called BioBIT (Biomedical Bert for ITalian). Finally, following a quality-over-quantity approach, two more checkpoints were developed with a new small corpus (200 MB, corresponding to 0.7% of the size of the PubMed corpus used in [7]); they are called $MedBIT_{R3}^+$ (implementing Mixout [29] to avoid Catastrophic Forgetting (CF)) and $MedBIT_{R12}^+$ (implementing Experience Replace [35] to avoid CF). CF is a phenomenon that causes neural networks to forget previously learned information upon being trained with new one [26], and it was taken into account while training these models by implementing Continual Learning techniques [27] in order to mitigate it [26, 28 - 31].

## Data collection and corpus construction

The main dataset used in this work, PsyNIT, was created starting from electronic medical reports collected by the IRCCS Istituto Centro San Giovanni di Dio Fatebenefratelli[4] hospital, located in Brescia, Italy. The documents contained various information about patients: demographic variables, medical history, results of tests and medical examinations, reports from medical exams, and more. Four sections of such documents were extracted:

- "Pharmacological history", usually a structured list of medications that the patient is taking and their dosages.

- "Remote pathologic history and active disease", usually a list of past and current relevant diseases.

---

[3] https://github.com/dbmdz/berts
[4] https://www.fatebenefratelli.it/

- "Cognitive proximate pathological history", typically unstructured, includes medical examinations the patient has undergone. It also includes information about the patient's personal life, such as marital status, daily habits, sleep disorders, and any relevant aspects of his/her behavior.

- "Psychological evaluation", typically unstructured, reports the result of (neuro)psychological examinations, together with comments from the attending physician.

PsyNIT was created from 100 medical reports. They were manually anonymized, removing personal patient data, physicians' references, dates, and locations. The anonymized documents were annotated by SF, psychologist and researcher with 10 years of experience, with the following classes of entities:

- "*DIAGNOSI E COMORBIDITÀ*" (779 examples, corresponding to 13.23% of the total dataset): Diagnosis and comorbidities, including medical concepts that encompass and identify a disease with a clinically classified definition. For our purposes, this class has been used to annotate both the main disease for which the medical report was written, and any other disease or medical condition, pre-existing or coexisting, from which the patient suffers. Examples are (Italian and translated): "*Neoplasia vescicale*" (bladder neoplasia), "*Ipoacusia*" (hearing loss), "*Ipofolatemia*" (hypopholatemia).

- "*SINTOMI COGNITIVI*" (2386 examples, corresponding to 40.52% of the total dataset): Cognitive symptoms, that reflect the individual's abilities in different cognitive domains. These are various aspects of high-level intellectual functioning, such as processing speed, reasoning, judgment, attention, memory, knowledge, decision-making, planning, language production and comprehension and visuospatial abilities [32]. In neuropsychiatric or cognitive disorders, various cognitive symptoms can be observed, showing the cognitive impairment of patients in different cognitive domains. Examples include: "*Anomia*" (anomie), "*Capacità introspettiva*" (introspective ability), "*Organizzazione e pianificazione visuospaziale*" (visuospatial organization and planning).

- "*SINTOMI NEUROPSICHIATRICI*" (707 examples, corresponding to 12.01% of the total dataset): Neuropsychiatric symptoms, that refer to a set of non-cognitive symptoms that occur in the majority of patients with dementia during the course of the disease [33]. These symptoms are referred to behavioral changes (such as mood disorders, anxiety, sleep problems, apathy, delusions, hallucinations), behavioral problems (like disinhibition, irritability or aggression), aberrant motor behavior and changes in eating behavior [34]. Examples include: "*Apatico*" (apathetic), "*Sintomi depressivi*" (depressive symptoms), "*Irritabile*" (irritable).

- "*TRATTAMENTO FARMACOLOGICO*" (162 examples, corresponding to 2.75% of the total dataset): Drug treatment, including any substance used to prevent or treat a medical problem, without dosage. Examples include: "*Madopar*", "*Urorec*".

- "*TEST*" (1854 examples, corresponding to 31.49% of the total dataset): Medical assessment, used to obtain an objective measure or information about a medical condition or disease. Examples include: "*EEG*" (ElectroEncephaloGram), "*MMSE*" (Mini-Mental State Examination), "*RM encefalo*" (brain magnetic resonance imaging).

The total number of examples of the above entities was 5888. Further details about annotation rules are presented in Supplementary notes, in section "Annotation rules", and more information related to PsyNIT in "Datasets analysis" section.

## Fine-tuning and evaluation metrics

PsyNIT was used to fine-tune four starting checkpoints: BaseBIT, BioBIT, MedBIT$_{R3}^{+}$, MedBIT$_{R12}^{+}$. The fine-tuning procedure has been repeated ten times for each model, initializing each run with a different random state, in order to minimize the effect of randomness and also to evaluate models' stability. The size of the test set was set at 10% of the dataset. The remaining data were split into a training set (80%) and a validation set (20%). BioBIT produced the highest average overall F1-score; the specific run that gave the closest F1-score to the average was chosen as reference model, and it will be now referred to as BioPsyNIT. The operations were carried out by means of a Python script built on well-known DL libraries (in particular Pytorch [35] and Transformers [36]). For this fine-tuning we implemented the same CF mitigation strategies adopted in [25]. The metrics used were Precision (P), Recall (R), and F1-score; we split the dataset in train and test subsets, and all the results reported have been measured on the test split (this also applies to the other experiments). Since classification is performed at token level, it is also important to note that the evaluation algorithm used the so-called IOB labelling format [37] (inside, outside, beginning) which means that each token is marked according to both its entity class and its position in the entity: "O" means that the token does not belong to any class (and therefore is not marked), "B" means that the token is the beginning of the entity, and "I" means the token is subsequent to a B or another I token. Details about fine-tuning, evaluation metrics, and IOB tagging are available in Supplementary notes, in the section "Fine-tuning and evaluation metrics".

## Multicenter experiments

Three external datasets have been crafted by as many hospitals: IRCCS Istituto Auxologico Italiano, in Milan, IRCCS Istituti Clinici Scientifici Maugeri and IRCCS Mondino Foundation, both in Pavia. The annotators were instructed in the same way as the annotator of PsyNIT. These datasets were used as independent test sets in Experiment 1, to evaluate the general applicability of BioPsyNIT, and in subsequent experiments to extend the training set. In this way, a total of four datasets were collected, allowing the design of several experiments:

1. One-vs-All (OvA): BioPsyNIT was tested with three external datasets. The goal was to see if it was flexible enough to work with unseen data, originating from different clinical centers.

2. Re-annotation: due to the low performance of the OvA approach (see section "Experiment 1"), a subset (approximately 20%) of the external datasets was re-annotated by the original PsyNIT annotator (version 2 of the datasets) to test whether a coherent annotation process could lead to better performance on independent datasets.

3. .1: Leave-One-Group-Out (LOGO): in this experiment, the starting checkpoints were fine-tuned on three datasets and tested on the fourth one, to see if using more data can improve performance on an independent dataset. Since every single dataset has different linguistic features, a model trained on multiple datasets should be able to generalize better [43] and thus performing well on unseen data (i.e., the dataset that was left out during the fine-tuning).

   .2: Low-resource fine-tuning: to further investigate the LOGO test results, low-resource fine-tuning of the models created in the previous experiment was performed with data from the left-out dataset. The goal was to test whether such models exploit the transfer learning approach and perform well even with a very small amount of unseen data (about 10% of the left-out dataset, which corresponds to a few hundred entities, while the model of the previous step was fine-tuned on thousands of entities).

4. Re-annotation plus Leave-One-Group-Out plus low-resource fine-tuning: in this experiment, we combined the findings from the previous experiments and simulated a scenario in which consistent, high-quality, independent datasets are combined with a LOGO plus the low-resource fine-tuning strategy to achieve an effective multicenter NER checkpoint.

5. Finally, we fine-tuned a model on the complete datasets combined, to create the model we are sharing with the community: MedPsyNIT.

Figure 1 shows the logical flow of the multicenter experiments. To be highlighted, every experiment that involved a fine-tuning phase implemented the CF mitigation strategies mentioned in the section "Methods – Pre-training".

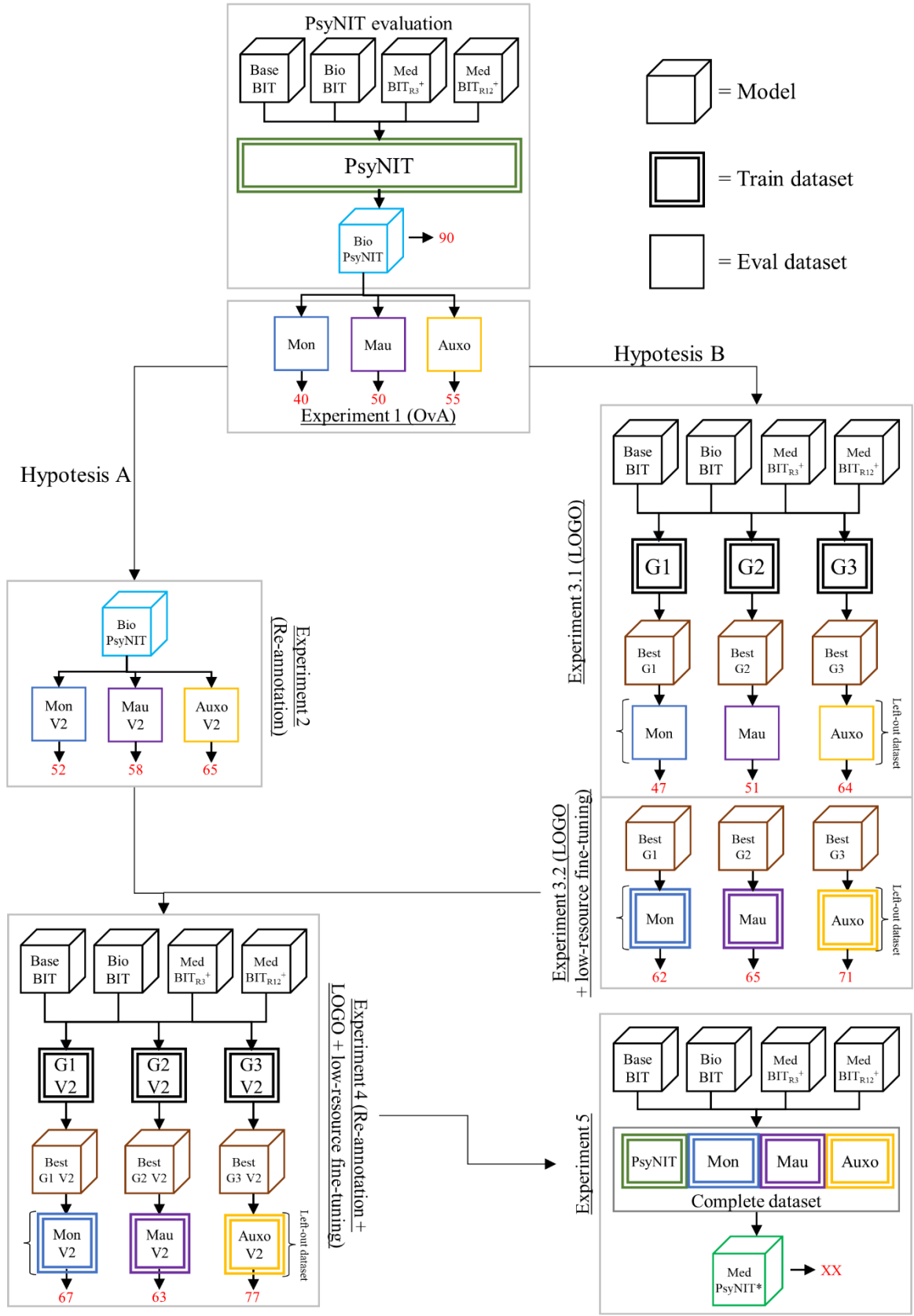

*Figure 1 - Workflow of multicenter experiments performed in this study. "Mon" = Mondino dataset, "Mau" = Maugeri dataset, "Auxo" = Auxologico dataset. First, we fine-tuned the four starting checkpoints with PsyNIT, creating BioPsyNIT. In Experiment 1, we tested it against 3 independent datasets; the average overall F1-score is reported in red for each dataset and for other*

*experiments as well. In Experiment 2 we tested BioPsyNIT against the re-annotated version of datasets, called V2. In Experiments 3.1 and 3.2, we applied a LOGO and LOGO followed by a few-shotlow-resource fine-tuning, respectively, on the four starting checkpoints, evaluating then the models with highest performance (Best G1, G2, G3). In Experiment 4, we combined the re-annotated datasets with the LOGO and few-shotlow-resource fine-tuning strategy, fine-tuning once again the starting checkpoints, demonstrating that this setup gives better peroformance than both Experiments 2 and in 2 out of 3 datasets. In Experiment 5 we fine-tuned the starting checkpoints to create MedPsyNIT, the model we are sharing with the community.*

# Results

## Datasets analysis

PsyNIT is composed in total by 5888 entities. The least represented class is *Drug treatment* (2.7% of the total dataset), while the most represented is *Cognitive symptoms* (40.5%). The external datasets, similarly to PsyNIT, were created from electronic medical reports of patients with neuropsychiatric disorders and annotated by physicians at the respective institutions; a complete overview is presented in Supplementary notes, in the section "Dataset analysis". They present different sizes, with the smallest being Auxologico (891 entities) and the largest being Maugeri (5949 entities). Because they came from different healthcare facilities, they had different writing styles and contained different entities (e.g., facilities might perform different types of tests or administer different drugs).

## PsyNIT evaluation

All the four selected checkpoints (i.e.: BaseBIT, BioBIT, MedBIT$_{R3}^+$, MedBIT$_{R12}^+$) have been fine-tuned with PsyNIT; the results are reported in Supplementary notes, in the section "PsyNIT evaluation". To check if statistical differences are present, we performed an ANOVA test [44] on the ten F1-scores of every model. The results show that there are no statistical differences between the score of the four models (p-value = 0.66). For subsequent experiments, we chose to use the checkpoint that gave the overall highest average F1-score, i.e., BioBIT (88.82% ± 2.24%). We picked the single run that gave the closest results to the average performance; its results are reported in Table 1. The overall F1-score is close to 90%, which is a high value compared to literature [20 – 23]. It is worth noticing that MedBIT$_{R3}^+$ and MedBIT$_{R12}^+$ achieved similar results (88.40% ± 2.32% and 88.18% ± 2.80%, respectively).

*Table 1 – Results of the PsyNIT fine-tuning process for the best performing checkpoint (BioBIT).*

| Class | F1 [%] | P [%] | R [%] |
|---|---|---|---|
| **DIAGNOSIS AND COMORBIDITIES** | 89.36 | 88.61 | 90.13 |
| **COGNITIVE SYMPTOMS** | 85.32 | 84.46 | 86.21 |
| **NEUROPSYCHIATRIC SYMPTOMS** | 83.57 | 78.07 | 89.90 |
| **DRUG TREATMENT** | 89.52 | 83.93 | 95.92 |
| **MEDICAL ASSESSMENT** | 98.31 | 96.67 | 100 |
| **OVERALL** | **89.53** | 87.38 | 91.80 |

# External dataset evaluation

### Experiment 1: One-vs-All

In the first experiment BioPsyNIT was evaluated with the three external datasets. The results are reported in Table 2. It is evident that performance on these datasets is lower than on PsyNIT; the overall F1-Score ranges from 40% to 55% (worst- and best-case scenario, respectively), while on PsyNIT it was 90%.

*Table 2 – Results of the evaluation process on Auxologico, Maugeri, and Modino datasets with the BioPsyNIT model.*

|  | Mondino | | | Maugeri | | | Auxologico | | |
|---|---|---|---|---|---|---|---|---|---|
| Class | F1 [%] | P [%] | R [%] | F1 [%] | P [%] | R [%] | F1 [%] | P [%] | R [%] |
| DIAGNOSIS AND COMORBIDITIES | 55.12 | 54.38 | 55.88 | 59.06 | 61.81 | 56.54 | 48.86 | 38.14 | 67.98 |
| COGNITIVE SYMPTOMS | 16.71 | 10.73 | 37.83 | 31.71 | 34.34 | 29.46 | 21.85 | 14.53 | 44.03 |
| NEUROPSYCHIATRIC SYMPTOMS | 32.97 | 25.95 | 45.18 | 52.85 | 58.33 | 48.32 | 37.84 | 27.85 | 58.99 |
| DRUG TREATMENT | 82.60 | 83.64 | 81.59 | 85.92 | 87.09 | 84.77 | 88.71 | 84.47 | 93.39 |
| MEDICAL ASSESSMENT | 39.17 | 43.54 | 35.60 | 43.79 | 52.20 | 37.71 | 49.90 | 44.35 | 57.05 |
| **OVERALL** | **40.17** | 33.54 | 50.07 | **50.76** | 55.36 | 46.87 | **55.33** | 44.96 | 71.91 |

### Experiment 2: Re-annotation

Experiment 1 showed that BioPsyNIT underperformed on external datasets. Since the corpora used to build the four datasets considered in this study are referred to patients with similar diagnoses, and subjected to similar tests, the hypothesis we formulated was that the poor performance is due to inconsistencies in the annotation process. To test this, a subset of about 20% of each external dataset was re-annotated from scratch by the PsyNIT annotator. The new datasets were then evaluated using BioPsyNIT; the results are shown in Table 3.

*Table 3 – Results of the evaluation process on a re-annotated subset (20%) of Auxologico, Maugeri, and Mondino datasets with the BioPsyNIT model.*

|  | Mondino | | | Maugeri | | | Auxologico | | |
|---|---|---|---|---|---|---|---|---|---|
| Class | F1 [%] | P [%] | R [%] | F1 [%] | P [%] | R [%] | F1 [%] | P [%] | R [%] |
| DIAGNOSIS AND COMORBIDITIES | 60.67 | 61.46 | 59.90 | 57.32 | 58.06 | 56.60 | 59.39 | 56.00 | 63.23 |
| COGNITIVE SYMPTOMS | 41.80 | 38.58 | 45.60 | 37.00 | 37.16 | 36.83 | 45.45 | 36.84 | 59.32 |
| NEUROPSYCHIATRIC SYMPTOMS | 57.85 | 47.30 | 74.47 | 68.21 | 58.86 | 81.10 | 45.78 | 35.19 | 65.52 |
| DRUG TREATMENT | 57.63 | 57.09 | 58.17 | 60.18 | 58.36 | 62.12 | 63.40 | 65.07 | 61.83 |
| MEDICAL ASSESSMENT | 87.50 | 84.85 | 90.32 | 84.73 | 87.59 | 82.04 | 86.36 | 81.62 | 91.69 |
| **OVERALL** | **52.50** | 49.41 | 56.00 | **58.01** | 57.15 | 58.89 | **65.10** | 59.67 | 71.61 |

### Experiment 3.1: Leave-One-Group-Out

To mitigate the low performance of the OvA, a LOGO approach was applied [45]. The number of groups was three, since PsyNIT was always included:

- Group 1 – fine-tuning datasets: PsyNIT, Auxologico, Maugeri; evaluation dataset: Mondino.

- Group 2 – fine-tuning datasets: PsyNIT, Auxologico, Mondino; evaluation dataset: Maugeri.

- Group 3 – fine-tuning datasets: PsyNIT, Mondino, Maugeri; evaluation dataset: Auxologico.

Although the BioBIT gave the best results on PsyNIT, we cannot assume that it will be the same in this experiment. Moreover, the ANOVA test showed no significant statistical differences between the results of the four starting checkpoints; for this reason, we used all of them for this experiment as well. Results have once again been averaged on ten random states, then the single model with performance closest to the average has been tested on each evaluation dataset. The highest overall F1-scores have been achieved by the MedBIT models on all the three groups (Table 4).

*Table 4 – Results of the LOGO process for the three groups on the fine-tuning test set and left-out set with the four starting checkpoints. Only the results for the best models are reported.*

| Class | Group 1 Best model (MedBIT$_{R3}^+$) | | Group 2 Best model (MedBIT$_{R12}^+$) | | Group 3 Best model (MedBIT$_{R12}^+$) | |
|---|---|---|---|---|---|---|
| | Fine-tuning test set | Left-out set | Fine-tuning test set | Left-out set | Fine-tuning test set | Left-out set |
| | F1 ± STD [%] | F1 [%] | F1 ± STD [%] | F1 [%] | F1 ± STD [%] | F1 [%] |
| DIAGNOSIS AND COMORBIDITIES | 81.54 ± 2.01 | 60.15 | 79.97 ± 2.78 | 69.67 | 78.5 ± 3.01 | 51.31 |
| COGNITIVE SYMPTOMS | 76.11 ± 2.59 | 21.69 | 72.62 ± 3.79 | 12.48 | 74.03 ± 2.05 | 38.94 |
| NEUROPSYCHIATRIC SYMPTOMS | 82.34 ± 3.84 | 42.41 | 71.53 ± 5.07 | 37.61 | 80.15 ± 2.52 | 54.04 |
| DRUG TREATMENT | 89.33 ± 2.80 | 49.50 | 90.80 ± 1.82 | 51.41 | 90.21 ± 2.38 | 57.55 |
| MEDICAL ASSESSMENT | 91.43 ± 4.44 | 85.89 | 92.13 ± 2.83 | 87.43 | 89.33 ± 5.20 | 91.14 |
| **OVERALL** | **83.52** ± 1.45 | **47.62** | **82.42** ± 1.38 | **51.35** | **81.96** ± 1.14 | **64.24** |

### Experiment 3.2: Low-resource fine-tuning

To better understand the results of Experiment 3.1, another test has been conducted. For every group, the best model has been further fine-tuned with 10% of the left-out dataset, corresponding to few hundred of entities. Figure 2 illustrates how the experiment has been carried out. The low-resource fine-tuning subset of the dataset is randomly selected; the experiment has been carried out with ten different random states, and average results are reported, in order to evaluate stability of the model on the test set. Models developed in Experiment 3.1 have been evaluated on the same test set, to compare them on the exact same dataset. Results are reported in Table 5.

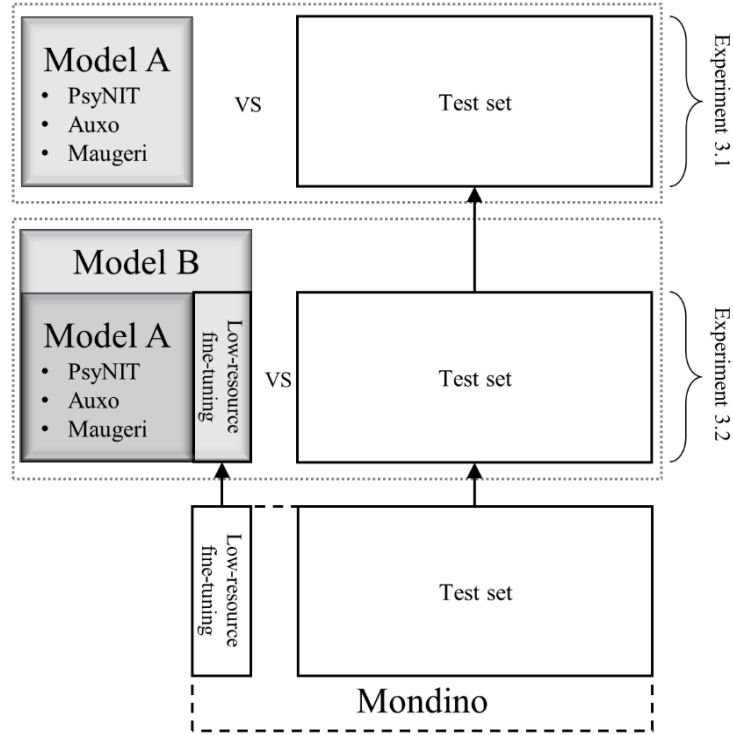

*Figure 2 - Design of Experiment 3.2. For illustration, we used Group 1, although it was also replicated for Group 2 and 3. Model A was fine-tuned with three datasets, while the left-out dataset was split into a few-shotlow-resource tuning and a test set. The first dataset was used to fine-tune model A into model B, while the second was used to evaluate the performance of both models.*

*Table 5 – Results of the LOGO process for the models A (trained in Experiment 3.1) and models B, fine-tuned with the low-resource approach.*

|  | Group 1, F1 ± STD [%] | | Group 2, F1 ± STD [%] | | Group 3, F1 ± STD [%] | |
| --- | --- | --- | --- | --- | --- | --- |
| **Class** | **Model A** | **Model B** | **Model A** | **Model B** | **Model A** | **Model B** |
| **DIAGNOSIS AND COMORBIDITIES** | 60.53 ± 0.35 | 62.39 ± 2.38 | 69.19 ± 0.41 | 68.25 ± 1.92 | 50.6 ± 2.00 | 61.38 ± 6.06 |
| **COGNITIVE SYMPTOMS** | 19.35 ± 0.38 | 41.17 ± 8.41 | 14.23 ± 0.25 | 51.26 ± 8.93 | 37.67 ± 3.38 | 47.08 ± 7.40 |
| **NEUROPSYCHIATRIC SYMPTOMS** | 40.09 ± 0.77 | 45.52 ± 3.68 | 43.37 ± 0.77 | 66.62 ± 7.29 | 56.24 ± 2.39 | 55.91 ± 2.18 |
| **DRUG TREATMENT** | 51.24 ± 0.42 | 64.37 ± 8.6 | 49.84 ± 0.46 | 63.85 ± 6.17 | 55.99 ± 1.42 | 68.46 ± 8.41 |
| **MEDICAL ASSESSMENT** | 85.23 ± 0.38 | 84.36 ± 0.84 | 85.71 ± 0.27 | 86.27 ± 0.74 | 91.35 ± 0.93 | 91.52 ± 1.23 |
| **OVERALL** | **46.57 ± 0.35** | **62.13 ± 3.99** | **50.87 ± 0.23** | **64.58 ± 5.29** | **64.11 ± 0.87** | **71.75 ± 3.46** |

## Experiment 4: Re-annotation + Leave-One-Group-Out + low-resource fine-tuning

Experiment 2 proved that a consistent annotation process is essential to increase performance between independent datasets. Experiment 3.2 showed that, even with inconsistently annotated datasets, the concatenation of LOGO strategy and low-resource fine-tuning allows to obtain relatively good performance. In this last experiment we combined these insights by performing the aforementioned training strategy with consistently annotated datasets. Results are reported in Table 6.

*Table 6 – Results of the LOGO + low-resource fine-tuning of the four starting checkpoints with re-annotated datasets. Only the results for the best models are reported.*

| Class | Group 1, F1 ± STD [%] Best model (MedBIT$_{R3}^+$) | Group 2, F1 ± STD [%] Best model (MedBIT$_{R12}^+$) | Group 3, F1 ± STD [%] Best model (MedBIT$_{R12}^+$) |
|---|---|---|---|
| DIAGNOSIS AND COMORBIDITIES | 70.1 ± 5.97 | 50.68 ± 10.06 | 70.12 ± 3.65 |
| COGNITIVE SYMPTOMS | 62.42 ± 7.46 | 46.39 ± 4.87 | 77.94 ± 4.96 |
| NEUROPSYCHIATRIC SYMPTOMS | 60.79 ± 3.11 | 65.86 ± 5.42 | 60.59 ± 9.60 |
| DRUG TREATMENT | 72.46 ± 4.48 | 73.88 ± 3.72 | 78.77 ± 4.59 |
| MEDICAL ASSESSMENT | 83.97 ± 7.53 | 84.53 ± 4.58 | 84.33 ± 3.98 |
| **OVERALL** | **66.73** ± 4.97 | **63.34** ± 1.95 | **77.19** ± 2.45 |

Experiment 5:

As a final experiment, we fine-tuned the four starting checkpoints on the complete dataset, i.e., the aggregation of PsyNIT and the three independent datasets. The starting checkpoint that yielded on average the highest F1-score on 10 seeds was MedBIT$_{R12}^+$. Since we wanted to share this model with the community, in this case we selected the single seed that performed the highest, i.e., F1-score 84.77%, Precision 83.16%, Recall 86.44%. The complete results are reported in Supplementary notes, in the section "Experiment 5". Moreover, we performed a further analysis on misclassifications; details can be found in the section "Error analysis" of Supplementary notes.

# Discussion

## Datasets analysis

The three external datasets have very different sizes. Nevertheless, this should not be a problem, since none of them will be used as a single dataset for fine-tuning. Looking at the total number of annotations, there is an imbalance between the entity classes. However, even the least represented class (*Medical assessment*, which is approximately 8% of the entities, as shown in "Dataset analysis" of Supplementary notes) has several hundred instances, so the model can learn to identify them correctly. The number of entities for the entire dataset is approximately 16000, making its size comparable to the literature. For comparison, here are some widely used biomedical NER datasets (e.g., used in the BioBERT article) with the number of annotations: BC5CDR [38], 15411; BC2GM [39], 20703; NCBI [40], 6881; and Species-800 [41], 3708.

# PsyNIT evaluation

PsyNIT was used to fine-tune four BERT-based checkpoints: BaseBIT, BioBIT, MedBIT$_{R3}^+$, and MedBIT$_{R12}^+$. The overall F1-score of BioPsyNIT is 90%, which represents a high value compared to the literature and even surpasses it in some cases [20 – 23]. The factors that led to this result are probably the quality of the starting checkpoints, the consistency of the dataset, and the annotation quality. The different entities showed different performances, ranging from 85 to 100%. Although *Drug treatment* is by far the least represented class with only 2.75% of the annotations, its F1-score is within the average of the other classes. This is likely due to the fact that the PsyNIT medical reports refer to a very narrow range of diagnoses, so the range of drugs listed in them is quite limited. This facilitated both the annotation process and the entity recognition algorithm and made disambiguation of drugs relatively easy. Another interesting thing to note is that for every class recall is higher than precision. This means that the model predicts more false positives than false negatives, thus tagging as entities tokens that are not. In a real-world practical application, such as an IE-based pipeline to convert clinical document into a structured database, high recall and low precision would mean that most of the important information is retrieved at the expense of some irrelevant data. Conversely, the opposite scenario would mean that most of the extracted information is relevant, but some important data is lost. The first option is preferable because filtering out irrelevant data requires less effort than re-elaborating reports to find missing information.

# External dataset evaluation

In this study we defined a multicenter Italian NER model. For this purpose, in addition to PsyNIT, three datasets were obtained from other clinical institutions: Mondino, Maugeri, and Auxologico. These datasets were derived from electronic clinical report forms written and annotated by clinicians at each center. They can be considered independent of each other, with the only commonality being that they refer to patients in the same diagnostic class. Several experiments were performed.

### Experiment 1: One-vs-All

The first experiment aimed to test the performance of BioPsyNIT on the three independent datasets. The results are relatively low, with overall F1-scores of 40%, 51%, and 55% for Mondino, Maugeri, and Auxologico, respectively. Two options were formulated to explain the significant performance gap with PsyNIT:

A. The datasets were annotated according to different rules, although we tried to make the annotation guidelines as clear as possible.

B. The datasets have some degree of intrinsic inconsistency, e.g., they cover broader medical topics, they have different grammatical structure, or they contain substantially different entities [42].

If option A is true, the low performance would be due to an inconsistent annotation process, which could be mitigated in two ways: first, by re-annotating the independent datasets with a consistent method; and second, by training the model with more data from different medical centers, thus achieving a better generalization ability [43]. Considering the goal of building a multicenter NER model, we deliberately chose not to pursue a strategy of a high inter-annotator agreement (IAA) at any cost. This would require too much effort to train each annotator iteratively and probably in several sessions to reach a sufficiently high IAA, which might be quite difficult in a real multicenter case scenario. Instead, our goal was to implement strategies to improve performance regardless of the consistency of annotations, while keeping the burden on the annotators as low as possible, so that other Italian centers wishing to take on and implement NER tasks in their own institutions could use our model. This said, accurate instructions for annotators in general remain an important point to provide for the development of an effective NER model; they can be found in the Supplementary notes, in the section "Annotation rules". Conversely, performance on inherently inconsistent datasets, as described in option B, will most likely remain poor, regardless of the quality of the annotations; while more data will allow for better performance due to higher generalizability.

Interestingly, F1-scores for the *Drug treatment* class are consistently above 80%, although overall performance for each dataset is low. This is likely due to the fact that, as with PsyNIT, disambiguation of drug entities is relatively easy due to similar suffixes in brand names, and therefore performance is high regardless of the dataset.

### Experiment 2: Re-annotation

To investigate the validity of option A, a second experiment was conducted, in which 20% of each external dataset was re-annotated by the same annotator who originally created PsyNIT, resulting in what is known as Version 2 (V2) of the datasets. The results show an improvement in performance for all datasets: Mondino increased from 40% to 52%, Maugeri from 51% to 58%, and Auxologico from 55% to 65%. This demonstrates that a consistent annotation process is needed to improve performance, but it remains relatively low, ranging from 52 to 65%.

### Experiment 3.1: Leave-One-Group-Out

To explore option B, Experiment 3.1 was conducted. It used a LOGO approach to test whether training the checkpoints with data from multiple centers would enable them to perform better on independent data (i.e., the left-out dataset during the fine-tuning phase). The results show that, despite the good performance during training (Overall F1-score for Group 1: 83%, Group 2: 82%, Group 3: 82%), the results on the external left-out datasets remain low, with an overall F1-score of 47%, 51%, and 64% for Mondino, Maugeri, and Auxologico, respectively. Interestingly, they outperformed the OvA approach even if their performance is relatively low. The lowest increase in F1-score compared with Experiment 1 is shown with the Maugeri evaluation set (+1%) and the highest with Auxologico (+9%), proving that the poor performances of OvA is due to the fine-tuning on a single dataset. Although the medical reports used to create datasets are narrative text, they tend to follow a specific grammatical pattern, peculiar to the physician who wrote them. For this reason, models fine-tuned on a single dataset are unlikely to have the generalization capability that would make them suitable for corpora with a different grammatical structure. Adding documents with different writing styles and possibly different medical terms and entities to the training set, allows the model to generalize more effectively and thus perform better on independent datasets.

Experiment 3.2: Low-resource fine-tuning

Further investigations were carried out on the LOGO approach. The models created in Experiment 3.1 were fine-tuned once more on a small number of files of the left-out group, implementing a low-resource learning approach. The goal was to demonstrate if the entire pre-training and fine-tuning processes executed so far, despite the low performance, enabled the checkpoints to perform NER effectively on unseen data even when trained on a relatively small number of entities. This would allow the models developed in this work to be used on independent datasets with minimal annotation effort and improving their usability and diffusion in healthcare facilities. For each group, we defined model A, which was fine-tuned as described in Experiment 3.1, and model B, which is model A plus the further low-resource fine-tuning. The left-out dataset is divided in two parts. The first part, representing 10% of the dataset (corresponding to a few hundred entities), was used for the low-resource fine-tuning; while the second part was used as a test set. The part of the dataset used for the low-resource fine-tuning was randomly selected; the experiment has been carried out 10 times, and average results are reported to evaluate stability of the model on the test set. The models created in Experiment 3.1 have been evaluated using the same test set to compare them to the exact same dataset (for this reason, results in the column "Eval dataset" from Table 4 and "Model A" column from Table 5 are different). The results show that B models consistently perform better than the corresponding A models. The lowest increase in F1-score compared to Experiment 3.1 is in Group 3 (+7.6%), where the starting F1-score was already close to 65%, while the highest is in Group 1 (+14.5%), where the performance reaches a score of 62%. This shows that the original checkpoints, fine-tuned with data from multiple centers, acquire a generalizability that allows them to be applied to new independent datasets, at the expense of a less resource-intensive annotation and training phase. This could pave the way to use a similar approach for institutions that cannot afford to create a large dataset, due to lack of data or resources for the annotation process.

Experiment 4: Re-annotation + Leave-One-Group-Out + low-resource fine-tuning

Experiment 4 was conducted combining the consistently re-annotated datasets created in Experiment 2 (option A) with the LOGO plus low-resource fine-tuning approach of Experiment 3.2 (option B). The objective was to demonstrate that the combination of these two approaches allows to obtain the best possible results by simulating a situation where consistently annotated datasets are used in addition to the training pipeline proposed. The results show that for two of the three external datasets (Mondino and Auxologico), the implemented approach outperformed both the Experiment 2 (re-annotated datasets) and Experiment 3.2 (LOGO plus low-resource fine-tuning). In particular, Mondino achieves an overall F1-score of 67%, and

Auxologico 77%. Conversely, performance for Maugeri dataset did not improve and dropped by about 1% compared to Experiment 3.2. This could be due to the fact that, despite our best effort to re-annotate in a consistent way and subsequent fine-tuning strategies, the grammatical structure of the dataset remains a key factor in determining performance. Comparing these results with Experiment 1 shows a significant performance increase: Mondino went from 40% to 67%, Maugeri from 50% to 63%, Auxologico from 55% to 77%. This proves the effectiveness of the implemented training pipeline, showing that the solution of the dilemma presented in Experiment 1 is the combination of both option A and option B. It is important to note that the goal of these experiments was not to achieve high performance per se, but rather to demonstrate the efficacy of the proposed training pipeline, which provided a significant increase in performance.

Experiment 5:

In this experiment we fine-tuned MedPsyNIT. We are making it available to the community, but, as demonstrated by previous experiments, it is unlikely to be ready for immediate use because it will first require a low-resource fine-tuning phase.

# Limitations and future work

The present work has limitations that can be overcome in future studies. The number of centers involved is relatively low, since annotating corpora is a time-consuming task, and so it is not trivial to find volunteers for it. Having demonstrated the effectiveness of the proposed pipeline, we could expand this approach to several other Italian medical centers. The second limitation consists of a pronounced imbalance in dataset sizes, with the smallest one (Auxologico, approximately 900 entities) being 85% smaller than the largest one (Maugeri, 6 thousand entities); this is partially connected to the annotation effort. For future works, we could set a threshold for the minimum and maximum number of entities of each class, thus also resolving the imbalance of single classes. Another limitation is the lack of an investigation that relates the syntactic structure and lexical content of datasets with their performance. In a future work we would like to identify linguistic features able to estimate the performance on the dataset, allowing us to give indications of the dataset quality even before the fine-tuning phase. Moreover, we could investigate the connection between the entities of the low-resource fine-tuning and the increase of performance, possibly by adding additional metrics other than F1-score. It is also fair to mention that several Large Language Models (LLMs) have been made publicly available in recent months, and these models can perform very well on many NLP tasks, including IE. However, unlike BERT-based models that can be deployed locally, the popular GPT-3.5 and GPT-4 require data to be transferred outside of hospitals, which has major ethical implications, considering the Health Insurance Portability and Accountability Act (HIPPA) and General Data Protection Regulation (GDPR[5]).

---

[5] https://gdpr.eu

In this study, we have shown that a consistent annotation process is crucial. Providing detailed instructions, with several examples, has not been enough. Therefore, as a further future step towards the creation of a consistent NER dataset, we propose the need for live training session where annotators could clarify doubts and agree on what is and what is not an entity.

The models we have created could be exploited to implement an IE tool that enables (semi-)automated data imputation for hospitals. To this day, these operations are often still performed by clinicians, with all the drawbacks that this implies, i.e., waste of time of highly specialized personnel and the prone-to-error nature of this task when performed on large texts. Moreover, NER is the first, basic step to perform more complex tasks (e.g., Relation Extraction). For this reason, the present work could pave the way for very task-specific IE tools.

# Conclusions

In this work, we created and shared with the community a native Italian neuropsychiatric NER dataset, with about six thousand entities divided into five categories, called PsyNIT. To our knowledge, this is the first publicly available Italian NER dataset[6] for neuropsychiatry. We used this dataset to fine-tune four Italian BERT checkpoints (one general-purpose, and three biomedical), creating BioPsyNIT. To check its general applicability, we tested it on three independent datasets (with a total of about 10 thousand entities) crafted by external Italian hospitals. We designed a set of experiments in order to mitigate annotation inconsistencies and to give the models the best possible generalization capabilities. The whole process highlighted a fundamental factor, namely that a multicenter model that can be used out-of-the-box is not effective and would likely provide low performance. However, a few hundred of high-quality, consistent examples, combined with a low-resource fine-tuning approach, can help to greatly enhance extraction quality. We believe that this evidence can be applied to other medical institutions and clinical settings, paving the way for the development of biomedical NER models in less-resourced languages.

# Environmental impact statement

The average computational cost we estimated for each fine-tuning run amounts to 0.75 GPU hours (on 4 models with 10 random states). Experiments have been carried out on the IRCCS Centro San Giovanni di Dio Fatebenefratelli high-performance computing environment, equipped with four A100 GPUs. Based on local rid carbon intensities[7] and hardware power consumptions, the calculation described in Luccioni et al. [46] results in a total of approximately 5.4 kgCO$_2$ eq produced, which is equivalent to 22 km driven by an average internal combustion engine car.

---

[6] https://huggingface.co/datasets/Neuroinformatica/PsyNIT
[7] https://www.isprambiente.gov.it/

# Acknowledgments

The present work was partially supported by "Ministero della Salute", IRCCS Research Program, Ricerca Corrente 2022-2023, Linea n. 1 "Utilizzo di strumenti di Intelligenza Artificiale (AI) per l'analisi dei disturbi psichici" and Ministry of Economy and Finance CCR-2017-23669078.

# Conflict of interest

None declared.

# Author contributions

CC: Conceptualization, Data curation, Formal analysis, Investigation, Methodology, Software, Validation, Writing – original draft, Writing – review & editing

TMB: Conceptualization, Methodology, Software, Writing – review & editing

SF: Methodology, Data curation, Writing – review & editing

EP: Conceptualization, Methodology, Writing – review & editing

FV: Methodology, Data curation

CF: Methodology, Data curation

MM: Methodology, Data curation

MCR: Methodology, Data curation

MC: Methodology, Data curation

AC: Methodology, Data curation

GB: Methodology, Data curation

RB: Conceptualization, Methodology, Supervision, Writing – review & editing

AR: Conceptualization, Data curation, Funding acquisition, Methodology, Resources, Supervision, Writing – original draft, Writing – review & editing

All authors reviewed the manuscript.

# Advancing Italian Biomedical Information Extraction with Transformers-based Models: Methodological Insights and a Practical Application

## Supplementary Notes

## Information Extraction example

Consider the following medical report:

*"Global cognitive state examination, assessed by the Mini-Mental State Examination (MMSE, raw score 22/30) shows good spatial orientation, deficit in temporal orientation. Global cognitive state examination, assessed by the Mini-Mental State Examination (MMSE, raw score 22/30) shows good spatial orientation, deficit in temporal orientation. The clock test (Clock drawing 5.69/10) detects visuo-spatial organization and planning skills at normal limits. Assessment of depressive symptoms (GDS 1/15) indicates absence of depressive symptomatology."*

Processing the report with a NER tool looking for *"Medical assessment"* entities should identify *"MMSE"*, *"Clock drawing"*, and *"GDS"*; the identified *"Medical assessment score"* entities should be *"22/30"*, *"5.69/10"*, and *"1/15"*. A RE tool should then link the identified entities with the respective scores: *MMSE → 22/30, Clock drawing → 5.69/10 GDS → 1/15*.

## Fine-tuning and evaluation metrics

The fine-tuning parameters are the following:

- Batch size = 10
- Learning rate = $3 \cdot 10^{-5}$
- Epochs = 50
- Weight decay = 0.01
- Warmup ratio = 0.02
- Layer-wise learning rate decay = 0.95
- Frozen layers = 3

The metric used are Precision (P), Recall (R), and F1-score, calculated as follows:

$$Precision = \frac{TP}{TP + FP}$$

$$Recall = \frac{TP}{TP + FN}$$

$$F1 = 2 \cdot \frac{P \cdot R}{P + R}$$

where:

- TP = Number of True Positives, i.e., tokens belonging to class X correctly identified as class X;
- FN = Number of False Positives, i.e., tokens belonging to another class, identified as class X;
- FN = Number of False Negatives, i.e., tokens belonging to class X identified as a different class.

As an example, consider the following sentence. For the sake of simplicity, let's assume that every word is a token (in a real case scenario, tokens are sub-words), and we have just two classes, *Medical assessment*, and *Drug treatment*:

"*Mister Rossi presents a MMSE of 22/30. At the present day, he is assuming Urorec twice a day, and Madopar once. He shows no symptoms of COVID-19.*"

The correct entities are:

- "*MMSE*" → *Medical assessment*
- "*Urorec*" → *Drug treatment*
- "*Madopar*" → *Drug treatment*

Let's say that the predicted entities are:

- "*Rossi*" → *Drug treatment*
- "*MMSE*" → *Medical assessment*
- "*Urorec*" → *Drug treatment*
- "*COVID-19*" → *Drug treatment*

*Medical assessment* class originally has one token, correctly identified ("*MMSE*"), and it presents no misclassifications, so it has one TP and no FPs nor FNs. *Drug treatment* class has originally two tokens, one of them is correctly identified ("*Urorec*"), thus producing a TP, but the other is not ("*Madopar*"), thus being a FN; moreover, there are two misclassifications ("*Rossi*" and "*COVID-19*"), producing two FPs. Table 1 and Table 2 summarize this example.

*Table 1 - Reference and predicted entities, number of True Positives, False Positives, and False Negatives, for the example sentence*

| Class | Entities | | | TP | FP | FN |
|---|---|---|---|---|---|---|
| | Ref | Pred | | | | |
| MEDICAL ASSESSMENT | MMSE | MMSE | TP | 1 | 0 | 0 |
| DRUG TREATMENT | Urorec | Urorec | TP | 1 | 2 | 1 |
| | Madopar | - | FN | | | |
| | - | Rossi | FP | | | |
| | - | COVID-19 | FP | | | |

*Table 2 - Formulas to calculate Precision, Recall, and F1-Score for the example sentence*

| Class | Precision | Recall | F1-Score |
|---|---|---|---|

| MEDICAL ASSESSMENT | $P = \dfrac{1}{1+0} = 1$ | $R = \dfrac{1}{1+0} = 1$ | $F1 = 2 \cdot \dfrac{P \cdot R}{P+R} = 2 \cdot \dfrac{1 \cdot 1}{1+1} = 1$ |
|---|---|---|---|
| DRUG TREATMENT | $P = \dfrac{1}{1+2} = 0.33$ | $R = \dfrac{1}{1+1} = 0.5$ | $F1 = 2 \cdot \dfrac{P \cdot R}{P+R} = 2 \cdot \dfrac{0.33 \cdot 0.55}{0.33+0.55} = 0.4$ |

So, for this example, we have that:

- *"Medical assessment"* has a F1-Score of 100%
- *"Drug treatment"* has a F1-Score of 40%

The overall F1-Score can be calculated as the weighted average of the two single F1-Scores, thus being:

$$F1 = \frac{100 \cdot 1 + 40 \cdot 2}{3} = 60\%$$

An example of IOB tagging, considering the following sentence: "*Gli esami evidenziano buon orientamento spaziale e temporale*" (meaning "Examinations show good spatial and temporal orientation"):

| | |
|---|---|
| Gli | O |
| esami | O |
| evidenziano | O |
| buon | O |
| orientamento | B-SINTOMI COGNITIVI |
| spaziale | I-SINTOMI COGNITIVI |
| e | O |
| temporale | B-SINTOMI COGNITIVI |

## Annotation rules

For annotations, we used the interface at the link provided in the footnote[1]. The basic instructions explained how to define entities classes, how to annotate words, and how to navigate in the document sections. The complete descriptions of entities classes can be found in the original paper. We remarked the importance of tagging entities separately, even if they are of the same type and consecutive. The interface, at the time of usage, did not allow to annotate overlapping entities. In this case, we gave instructions to annotate the most inclusive entity fully, and the others partially. In general, our instructions were to be as concise as possible when going to highlight an entity. As a "rule of thumb", one should consider anything related to the entity (adjectives, complements, ...) that is not essential for its identification not part of the entity, that in practical terms means "when in doubt be as restrictive as possible". Some class specific rules:

- For *"Medical assessment"* entities, only the name of the assessment had to be tagged, omitting its score;

- For *"Drug treatment"* entities, only the name of the treatment has to be tagged, omitting its dosage;

---

[1] https://tecoholic.github.io/ner-annotator/

# PsyNIT entities

Table 3 reports the total number of annotations in PsyNIT.

*Table 3 - Total number of entities and percentage for every class for PsyNIT*

| Class | Numerosity # | Percentage [%] |
|---|---|---|
| DIAGNOSIS AND COMORBIDITIES | 779 | 13.23 |
| COGNITIVE SYMPTOMS | 2386 | 40.52 |
| NEUROPSYCHIATRIC SYMPTOMS | 707 | 12.01 |
| DRUG TREATMENT | 162 | 2.75 |
| MEDICAL ASSESSMENT | 1854 | 31.49 |
| OVERALL | 5888 | 100 |

# External datasets entities

The three external datasets have been created starting from clinical document of the respective hospitals. Here is reported the number of documents used for every center:

- Mondino = 60

- Maugeri = 60

- Auxologico = 20

Table 4 reports the number of entities, divided by class, for every dataset.

*Table 4 – Number of entities for every external dataset, divided by class*

| | Mondino | | Maugeri | | Auxologico | |
|---|---|---|---|---|---|---|
| Class | # | % | # | % | # | % |
| DIAGNOSIS AND COMORBIDITIES | 586 | 17.92 | 626 | 10.52 | 167 | 18.74 |
| COGNITIVE SYMPTOMS | 726 | 22.20 | 2471 | 41.54 | 141 | 15.82 |
| NEUROPSYCHIATRIC SYMPTOMS | 360 | 11.01 | 861 | 14.47 | 90 | 10.10 |
| DRUG TREATMENT | 365 | 11.16 | 515 | 8.66 | 266 | 29.85 |
| MEDICAL ASSESSMENT | 1233 | 37.71 | 1476 | 24.81 | 227 | 25.48 |
| OVERALL | 3270 | 100 | 5949 | 100 | 891 | 100 |

The number of documents for Auxologico, and consequently the number of annotations, is sensibly lower compared to the other datasets, because annotating documents is a time-consuming task, and not all clinicians could dedicate the same time to this activity. However, we decided to include the Auxologico dataset as well, regardless of its size, to prove the general applicability of PsyNIT even on small datasets.

Table 5 reports, for every external dataset, the following numbers:

- Common entities: the number of common entities between PsyNIT and the dataset;

- Common annotations: the number of annotations corresponding to the common entities, as absolute and relative number. For example, for Mondino dataset, the class "Drug treatment" shares 48 type of entities with PsyNIT, corresponding to 213 annotations. These are the 28.48% of the total annotations of "Drug treatment" class for Mondino dataset

*Table 5 - Common entities and annotations between PsyNIT and external datasets*

| Class | Mondino | | | Maugeri | | | Auxologico | | | TOTAL | |
|---|---|---|---|---|---|---|---|---|---|---|---|
| | Common entities | Common annotations | | Common entities | Common annotations | | Common entities | Common annotations | | Average | Stdev |
| DIAGNOSIS AND COMORBIDITIES | 49 | 159 | 27,13% | 63 | 179 | 28,59% | 28 | 64 | 38,32% | 31,35% | 6,08% |
| COGNITIVE SYMPTOMS | 60 | 527 | 16,40% | 45 | 409 | 16,55% | 7 | 9 | 6,38% | 13,11% | 5,83% |
| NEUROPSYCHIATRIC SYMPTOMS | 40 | 280 | 24,24% | 28 | 196 | 22,76% | 13 | 25 | 27,78% | 24,93% | 2,58% |
| DRUG TREATMENT | 48 | 213 | 28,48% | 39 | 152 | 29,51% | 23 | 84 | 31,58% | 29,86% | 1,58% |
| MEDICAL ASSESSMENT | 20 | 531 | 20,42% | 15 | 287 | 19,44% | 8 | 82 | 36,12% | 25,33% | 9,36% |
| OVERALL | 217 | 1710 | 20,59% | 190 | 1223 | 20,56% | 79 | 264 | 29,63% | 23,59% | 5,23% |

# Inter Annotator Agreement

When dealing with datasets that involve annotation processes, in order to create a high-quality dataset, it is important for annotators to be consistent with each other. In practical terms this means that corpora have to be labeled coherently: in a NER scenario, if a group of words is marked as entity by an annotator in one corpus, the same group of words should be labeled with the same entity class in all other corpora. A measure that quantifies the level of agreement between annotators is the Inter-Annotator Agreement (IAA). IAA can be calculated in several ways, e.g., assuming that one annotator is the gold reference and calculating the F1-score of other annotators. IAA can be classified according to its value:

- Poor Agreement (0.0 - 0.2): there are significant discrepancies in annotation;

- Fair Agreement (0.2 - 0.4): there are substantial differences in annotations;

- Moderate Agreement (0.4 - 0.6): annotators show reasonable agreement;

- Good Agreement (0.6 - 0.8): the annotations are reliable for most practical purposes;

- Excellent Agreement (0.8 - 1.0): this indicates very consistent annotations.

Considering the goal of building a multicenter NER model, we deliberately chose not to pursue a strategy of a high inter-annotator agreement (IAA) at any cost, because this would require a large effort. Instead, our goal was to implement strategies to improve performance regardless of the consistency of annotations, while keeping the burden on the annotators as low as possible. Nevertheless, we calculated a partial IAA by exploiting the original annotation of the external datasets and the re-annotated 20% by the PsyNIT original annotator (see "External dataset evaluation" section, at paragraph "Experiment 2"). Table 6 reports IAA results.

*Table 6 - Total number of entities and percentage for every class for PsyNIT*

| Class | Mondino | Maugeri | Auxologico | Average |
|---|---|---|---|---|
| MEDICAL ASSESSMENT | 0.34 | 0.46 | 0.45 | 0.42 ± 0.07 |
| COGNITIVE SYMPTOMS | 0.31 | 0.43 | 0.09 | 0.28 ± 0.17 |
| NEUROPSYCHIATRIC SYMPTOMS | 0.65 | 0.50 | 0.46 | 0.54 ± 0.10 |
| DIAGNOSIS AND COMORBIDITIES | 0.56 | 0.54 | 0.53 | 0.54 ± 0.02 |
| DRUG TREATMENT | 0.83 | 0.93 | 0.69 | 0.82 ± 0.12 |
| OVERALL | 0.54 ± 0.22 | 0.57 ± 0.20 | 0.44 ± 0.22 | 0.52 ± 0.21 |

Results show that:

- IAA for "Drug treatment" is good/excellent. This is somehow expected because, as mentioned in the "PsyNIT evaluation" section of the article, the small range of diagnoses narrows the list of drugs in the corpora, facilitating the annotation process. Indeed, the class "Drug treatment" is the one that presents the highest scores in several experiments;

- IAA for "Cognitive symptoms" is the lowest (poor/fair). This reflects on the training process, making it the class with lowest F1-scores several times;

- IAA for other classes ("Medical assessment" = 0.42, "Neuropsychiatric symptoms" = 0.54, "Diagnosis and comorbidities" = 0.54) is moderate.

# Experiment 5

Results of experiment 5 are reported in Table 7.

*Table 7 – Overall results for the fine-tuning process with four different checkpoints on the Compelte dataset.*

| | BaseBIT | BioBIT | MedBIT$_{R3}^+$ | MedBIT$_{R12}^+$ |
|---|---|---|---|---|
| Class | F1 ± STD [%] | F1 ± STD [%] | F1 ± STD [%] | F1 ± STD [%] |
| DIAGNOSIS AND COMORBIDITIES | 75.09 ± 3.14 | 75.92 ± 3.15 | 75.92 ± 3.02 | 76.12 ± 2.52 |
| COGNITIVE SYMPTOMS | 71.37 ± 3.15 | 73.21 ± 3.23 | 72.35 ± 4.12 | 73.01 ± 3.95 |
| NEUROPSYCHIATRIC SYMPTOMS | 77.62 ± 3.38 | 77.2 ± 3.96 | 77.28 ± 4.56 | 77.78 ± 4.47 |
| DRUG TREATMENT | 88.63 ± 2.4 | 89.01 ± 2.15 | 89.35 ± 2.15 | 89.18 ± 1.86 |
| MEDICAL ASSESSMENT | 89.11 ± 5.28 | 89.01 ± 4.47 | 89.52 ± 4.1 | 89.59 ± 4.88 |
| **OVERALL** | **80.23 ± 1.82** | **80.94 ± 2.18** | **80.87 ± 2.34** | **81.14 ± 2.13** |

# Error analysis

The reference metrics in all experiments has been F1-score, calculated as harmonic mean of precision and recall. Thus, it is fundamental to identify true positives, false positives, and false negatives. Nevertheless, one could argue that not all errors have the same impact on a NER pipeline, e.g., tagging a *B-ClassX* token as *I-ClassX* is a less impactful mistake than tagging it as *O*. For this reason, alongside to false positives and false negatives (also known as type 1 and type 2 errors, respectively), two other error classes have been defined:

- Type 3 errors: token identified, but wrong semantic type (e.g., *B-TEST* identified as *B-DIAGNOSI E COMORBIDITÀ*)

- Type 4 errors: token identified, correct semantic type, but IOB tag is wrong (e.g., *B-TEST* identified as *I-TEST*)

This evaluation has been performed on Total dataset, and results can be seen in Table 8.

*Table 8 – Token classification errors divided by class*

| Class | # | % |
|---|---|---|
| **False Positives (Type 1)** | 526 | 47.30 |
| **False Negatives (Type 2)** | 343 | 30.85 |
| **Type 3** | 28 | 2.52 |
| **Type 4** | 215 | 19.33 |
| **TOTAL** | 1112 | 100 |

Results show that the most frequent errors are false positives, which represent ~47% of the total. This is expected, since the overall Precision (83%) is lower than the overall Recall (86%). Interesting to notice, type 3 errors are only the 2.5% of the total, meaning that it is very infrequent for the model to misclassify between entities classes. Type 4 errors are common (~19%), meaning that 1 error out of 5 is an incorrect IOB classification. This could have an impact on the overall performance, depending on how the word is reconstructed starting from the tokens that compose it.